\pdfoutput=1

\documentclass[bold,a4paper,10pt]{paper}

\usepackage[margin=1in]{geometry}
\usepackage[utf8]{inputenc}
\usepackage[T1]{fontenc}
\usepackage{hyperref}
\usepackage{booktabs}
\usepackage{amsfonts}
\usepackage{amsmath}
\usepackage[final]{microtype}
\usepackage[dvipsnames]{xcolor}
\usepackage{graphicx}
\usepackage{natbib}
\usepackage{subcaption}
\usepackage{multirow}

\makeatletter\let\NAT@force@numbers\relax\makeatother

\title{Toloka Visual Question Answering Benchmark}

\author{
  \textbf{Dmitry Ustalov}$^\P$\thanks{This work was done while the author was with Toloka.} \\
  \texttt{dmitry.ustalov@jetbrains.com} \\
  \and
  \textbf{Nikita Pavlichenko}$^\dag$ \\
  \texttt{pavlichenko@toloka.ai} \\
  \and
  \textbf{Sergey Koshelev}$^\dag$ \\
  \texttt{korzg@toloka.ai} \\
  \and
  \textbf{Daniil Likhobaba}$^\dag$ \\
  \texttt{likhobaba-dp@toloka.ai} \\
  \and
  \textbf{Alisa Smirnova}$^\ddag$ \\
  \texttt{zero@toloka.ai}
}

\institution{
  \normalfont{}$^\P$ JetBrains \\
  \normalfont{}Belgrade, 11070 Serbia \\
  \and
  \normalfont{}$^\dag$ Toloka \\
  \normalfont{}Belgrade, 11000 Serbia \\
  \and
  \normalfont{}$^\ddag$ Toloka \\
  \normalfont{}Lucerne, 6005 Switzerland \\
}

\begin{document}

\maketitle

\begin{abstract}
In this paper, we present \emph{Toloka Visual Question Answering}, a new crowdsourced dataset allowing comparing performance of machine learning systems against human level of expertise in the grounding visual question answering task. In this task, \emph{given an image and a textual question, one has to draw the bounding box around the object correctly responding to that question}. Every image-question pair contains the response, with only one correct response per image. Our dataset contains 45,199 pairs of images and questions in English, provided with ground truth bounding boxes, split into train and two test subsets. Besides describing the dataset and releasing it under a CC~BY license, we conducted a series of experiments on open source zero-shot baseline models and organized a multi-phase competition at WSDM~Cup that attracted 48 participants worldwide. However, by the time of paper submission, no machine learning model outperformed the non-expert crowdsourcing baseline according to the intersection over union evaluation score.
\end{abstract}

\begin{figure}[htbp]
  \centering
  \subcaptionbox{What do we use to support the immune system and get vitamin C?}{\includegraphics[width=.31\linewidth]{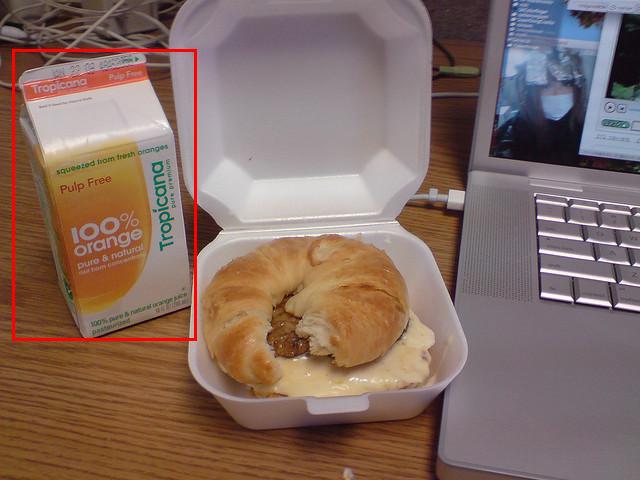}}
  \hfill
  \subcaptionbox{What do people use for cutting?}{\includegraphics[width=.31\linewidth]{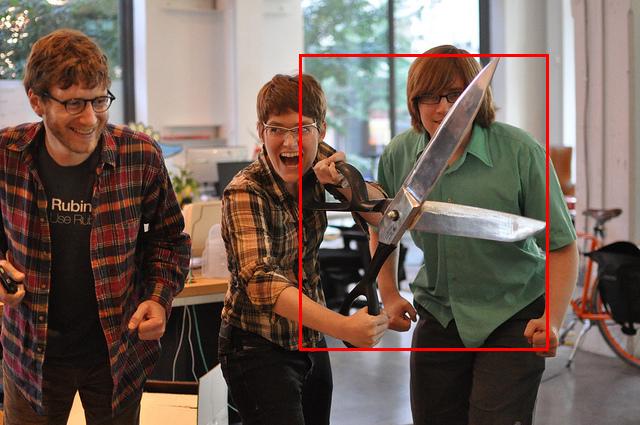}}
  \hfill
  \subcaptionbox{What do you use to hit the ball?}{\includegraphics[width=.31\linewidth]{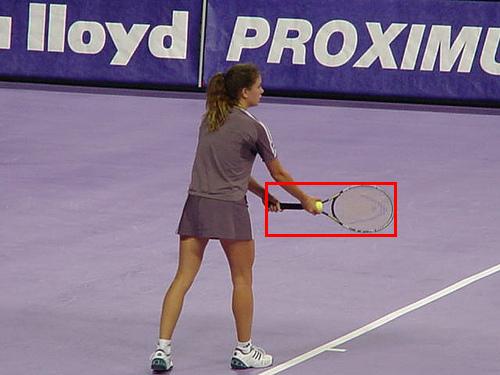}}
  \caption{\label{fig:teaser}\textbf{Given an image and a textual question, draw a bounding box containing the correct answer to the question.} Above is a sample of three image-question pairs from the training subset of our dataset. Every image contains the response, with only one correct response per image. Bounding boxes are drawn for illustrative purposes only; they are not parts of images in our dataset but are available as ground truth for all images. All images are from the MS~COCO dataset \citep{Lin:14} under the same license.}
\end{figure}

\clearpage

\section{Introduction}

Recently, prominent multi-modal deep learning models such as CLIP \citep{Radford:21} and DALL-E \citep{Ramesh:21} have demonstrated remarkable performance in demanding tasks such as text-image similarity measurement and text-to-image generation, respectively. Concurrently, modern machine learning methods have achieved superhuman results on challenging multi-task benchmarks like SuperGLUE \citep{Wang:19} and VLUE \citep{Zhou:22}. However, most of these benchmarks incorporate a combination of well-known tasks with limited modality. In this study, we enhance the level of difficulty for machine learning methods by introducing the \emph{Toloka Visual Question Answering}, an open-source multi-modal dataset designed to evaluate artificial intelligence systems. We provide a comprehensive description of the benchmark, outline our crowdsourcing pipeline for data collection, and present the performance of current pre-trained and fine-tuned models in tackling the challenging problem of grounding visual question answering.

Our task is formulated as follows. Given an image and an English textual question, the objective is to draw a bounding box around the object that provides the correct response to the question (Figure~\ref{fig:teaser}). For instance, in a photograph of a bathroom, a question like ``Where do I wash my hands?'' would require selecting the sink as the answer. Successfully solving this task necessitates the non-trivial integration of visual, textual, and commonsense information. We assert that our approach, which employs free-form, open-ended textual questions paired with bounding boxes as answers, presents a fair challenge for contemporary multi-modal models.

The remainder of this paper is organized as follows: Section~\ref{sec:related} provides an overview of related work, Section~\ref{sec:dataset} introduces our grounding visual question answering dataset, Section~\ref{sec:annotation} describes the annotation pipeline employed to create the dataset, Section~\ref{sec:metrics} defines the evaluation metrics and baselines used for assessment, Section~\ref{sec:evaluation} presents the evaluation results for publicly-available models and submissions in our competition, Section~\ref{sec:errors} conducts an error analysis of both human and machine performance on our dataset, and finally, Section~\ref{sec:conclusion} outlines the limitations of our work and concludes with final remarks.

\section{\label{sec:related}Related Work}

In recent years, the scientific community has made significant progress in the development of diverse datasets containing multi-modal data, enabling numerous applications at the intersection of natural language processing and computer vision. One prominent application in this domain is visual question answering (VQA)~\citep{Antol:15}, where models are tasked with providing textual responses based on image-question pairs, often involving commonsense knowledge. Several datasets have been created to facilitate research in VQA, such as GQA~\citep{Hudson:19}, CLEVR~\citep{CLEVR}, and VQA v2~\citep{VQAv2}, which leverage MS COCO\footnote{\url{https://cocodataset.org/}} images \citep{Lin:14} (which is also the case for our dataset).

However, the conventional VQA paradigm assumes that the output should be in textual form. In contrast, visual grounding task requires to find a region of an image referring to a textual description of an image. The RefClef, RefCOCO, RefCOCO+~\citep{RefCOCO}, and \textsc{GrIT} \citep{Kosmos-2} datasets are examples that highlight the challenges of this visual grounding task. Our work lies at the intersection of these tasks, requiring both natural language understanding to process the question and the ability to comprehend the visual scene to detect relevant objects. We frame the problem as a \emph{grounding visual question answering task}, where the model must output an object identified by a bounding box as the answer to the question. It is important to note that our problem cannot be easily reduced to the standard text-only question answering or detection based solely on textual prompts, as the answer to the question depends on the content of the image. In grounding VQA task \citep{Zhu:16,Qiao:21,Chen_2022_CVPR} one predicts the region in the image used to arrive at the answer but not necessarily the answer itself. We guarantee that the answer is always present, yet we firmly believe that our proposed setup presents a formidable challenge for modern multi-modal models.

\section{\label{sec:dataset}Dataset Description}

Our dataset is comprised of the images associated with textual questions (Figure~\ref{fig:teaser}). One entry (instance) in our dataset is a question-image pair labeled with the ground truth coordinates of a bounding box containing the object answering the given question. We guarantee that in most cases each image contains one and only one correct response to the given question. The images were obtained from a subset of the Microsoft Common Objects in Context, MS~COCO, dataset \citep{Lin:14} that was licensed under the Creative Commons Attribution (CC~BY) license.

\begin{table}[t]
\centering
\caption{\label{tab:dataset}Descriptive statistics of our dataset. There are 45,199 instances in total. Image and bounding box dimensions are in pixels. Question lengths are in characters including spaces. All numbers are 95\% confidence intervals, except the number of instances.}
\begin{tabular}{lrccccc}\toprule
\multirow{2}{*}{\textbf{Subset}} & \multirow{2}{*}{\textbf{\# of instances}} & \multicolumn{2}{c}{\textbf{Image}} & \multicolumn{2}{c}{\textbf{Bounding Box}} & \textbf{Question} \\
 &  & \textbf{Width} & \textbf{Height} & \textbf{Width} & \textbf{Height} & \textbf{Length} \\ \midrule
train & 38,990 & $(578, 581)$ & $(486, 489)$ & $(103, 105)$ & $(97, 99)$ & $(36, 38)$ \\
public test & 1,705 & $(573, 583)$ & $(483, 493)$ & $(91, 100)$ & $(88, 97)$ & $(36, 39)$ \\
private test & 4,504 & $(578, 584)$ & $(480, 487)$ & $(95, 101)$ & $(89, 94)$ & $(36, 38)$ \\ \bottomrule
\end{tabular}
\end{table}

\begin{figure}[t]
  \centering
  \subcaptionbox{\label{fig:heatmap}Heat map of bounding boxes in the private test subset of our dataset. Heat maps of other subsets look similarly.}{\includegraphics[width=.45\linewidth]{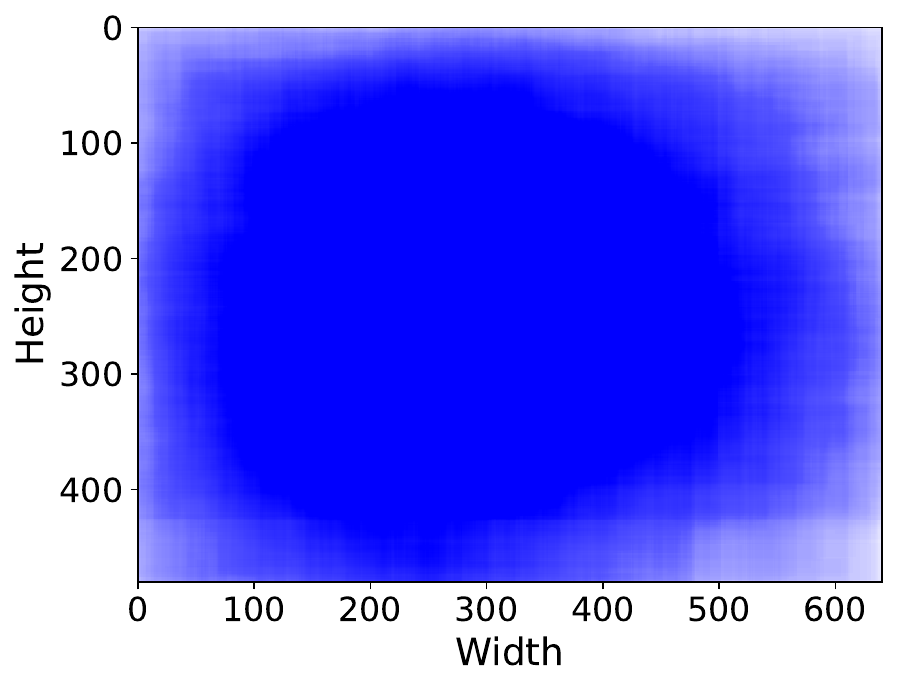}}
  \hfill
  \subcaptionbox{\label{fig:pie}Diagram of the 30 most common types of objects making 26\% of the objects in our dataset.}
  {\includegraphics[width=.45\linewidth]{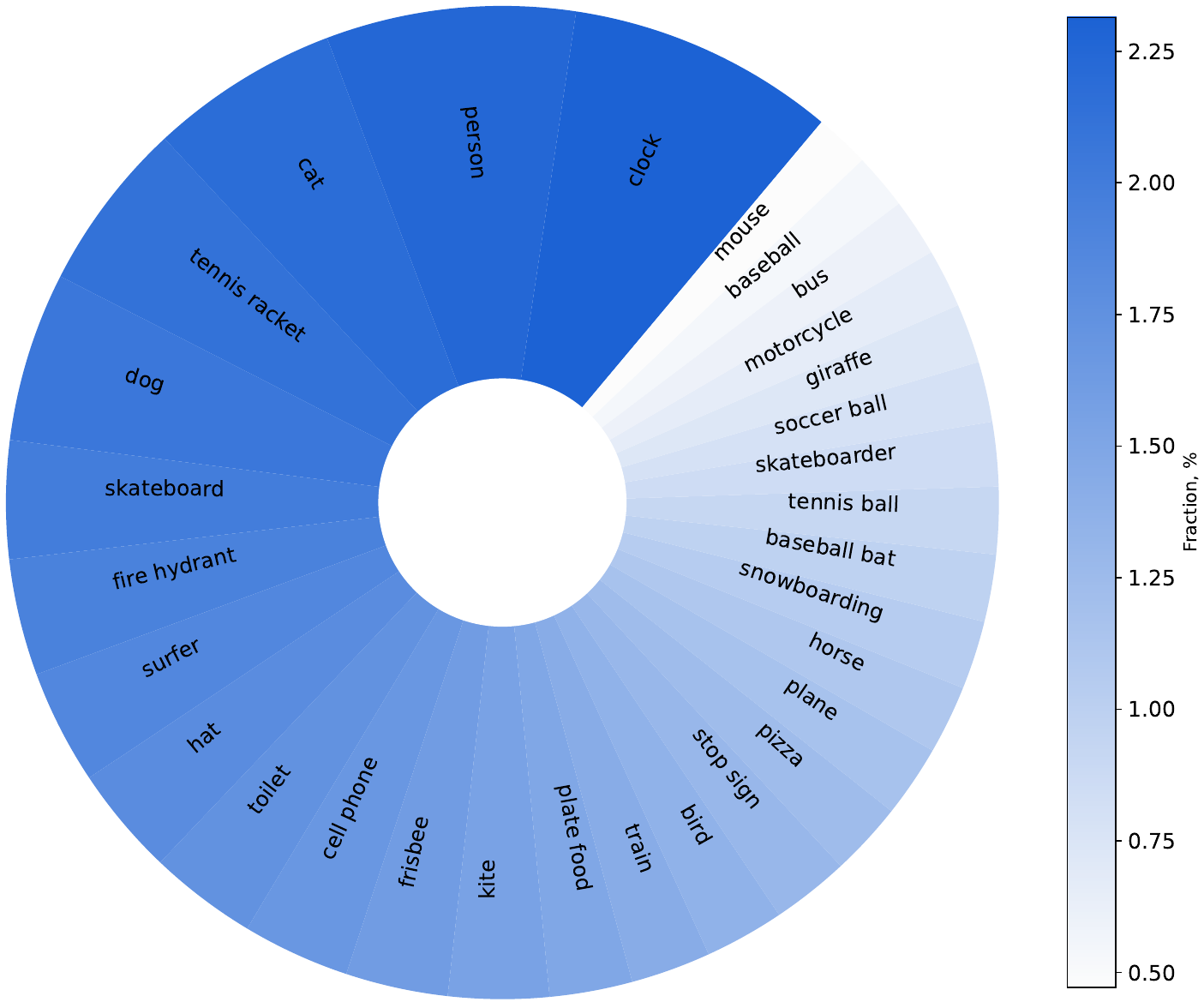}}
  \caption{Visual analysis of the ground truth bounding boxes in our dataset.}
\end{figure}

Our dataset consists of 45,199 instances, which are divided into three subsets as shown in Table~\ref{tab:dataset}: \emph{train} (38,990 instances), \emph{public test} (1,705 instances), and \emph{private test} (4,504 instances). The names of these subsets correspond to the different phases of the competition that we organized (refer to Section~\ref{sec:evaluation} for further information). Since the public release of the entire dataset the \emph{train} subset can be used as the training set, the \emph{public test} subset as the validation set, and \emph{private test} as the test set. The dataset is provided in the form of textual files in comma-separated value (CSV) format, containing the following information: (a) URL of an image on a public content delivery network, (b) question in English, (c) image width and height, and (d) bounding box coordinates (left, top, right, bottom).

We made our complete dataset, along with the baselines and ground truth bounding boxes, publicly available on various platforms to encourage research and development of multi-modal question answering models. The dataset can be accessed on Zenodo,\footnote{\url{https://doi.org/10.5281/zenodo.7057740}} Hugging Face Hub,\footnote{\url{https://huggingface.co/datasets/toloka/WSDMCup2023}} Kaggle,\footnote{\url{https://www.kaggle.com/datasets/dustalov/toloka-wsdm-cup-2023-vqa}} and GitHub.\footnote{\url{https://github.com/Toloka/WSDMCup2023}} It was released under the same CC~BY license as the MS~COCO subset we used. To ensure the integrity of the dataset and address potential concerns regarding dataset split-view poisoning \citep{carlini2023poisoning}, we computed and uploaded SHA-256 hashes for the images and electronically signed the repository commits that contain our data files. Additionally, we have uploaded all the images to Zenodo and Kaggle to mitigate any potential unavailability issues with the Azure content delivery network that we utilized to store the images.

Our dataset has an equal proportion of images and questions to allow capturing different parts of different images, making it useful for training both visual and textual aspects of the model. Descriptive statistics of our dataset are presented in Table~\ref{tab:dataset}. It is evident that the subsets share a similar structure, with the majority of bounding boxes located near the centers of the images (Figure~\ref{fig:heatmap}). Furthermore, we extracted portions of images from the dataset, confined within the bounding boxes, and captioned them using BLIP-2~\citep{blip2}. This process resulted in textual descriptions of the selected objects. By clustering these descriptions in the \emph{test private} subset, we found that 72\% of them formed 65 distinct clusters, while 28\% of objects did not belong to any cluster, demonstrating the diversity and non-triviality of our dataset. The 30 most common types of objects enclosed in bounding boxes are illustrated in Figure~\ref{fig:pie}.

To further analyze the questions in our dataset, we sampled 100 random questions from the private test subset. Then, three authors of the paper manually annotated whether it is possible to answer the given question without seeing an image. We aggregated the annotations by majority vote; the inter-annotator agreement was high as indicated by Krippendorff's $\alpha = 0.87$ \citep{Krippendorff:18}. Our evaluation showed that only 56\% of questions were answerable without seeing an image. That is, there was a specific answer to these questions (e.g., ``What does the baseball player use to hit the ball?'') while the rest 44\% of questions have multiple answers (e.g., ``What can be used to eat food with?'') or the question is specifically about the image (e.g., ``What is the person riding?''). Thus, our analysis shows that almost a half of the questions cannot be answered without seeing an image.

\section{\label{sec:annotation}Annotation Methodology}

\begin{figure}[t]
  % https://docs.google.com/drawings/d/1xAlfzQMwOeM9FaPxyRLYsJVeTRvEh4Q8k3pCO353VIA/edit?usp=sharing
  \centering
  \includegraphics[width=.8\linewidth]{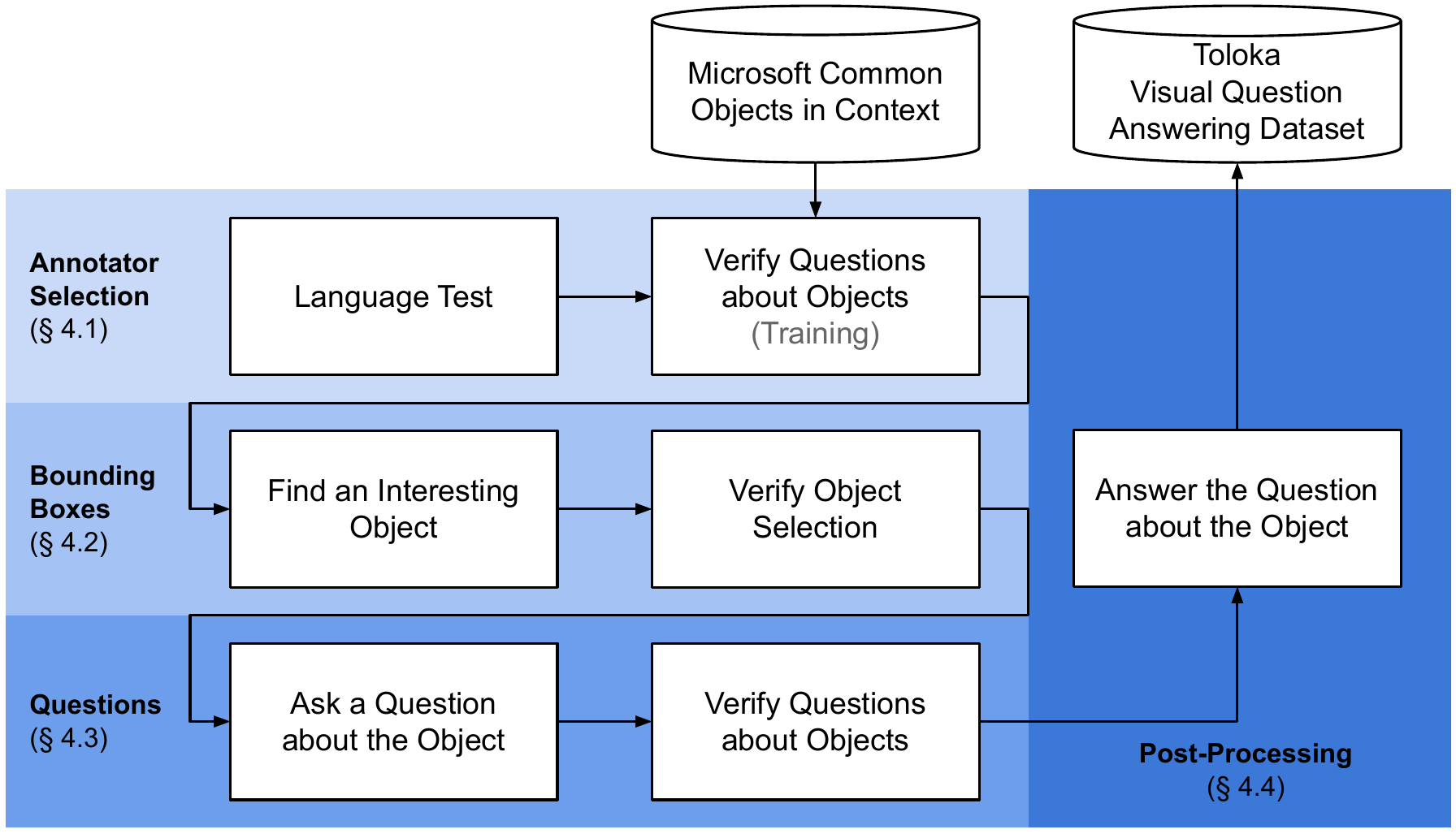}
  \caption{\label{fig:pipeline}A diagram of our annotation pipeline. First, we perform an \emph{annotator selection} (Section~\ref{sub:annotators}). Then, we ask the annotators to draw \emph{bounding boxes} around interesting objects (Section~\ref{sub:boxes}) and to \emph{compose questions} (Section~\ref{sub:questions}). Finally, during \emph{post-processing}, we ask the annotators to answer the composed image-question pairs (Section~\ref{sub:processing}).}
\end{figure}

We performed all annotations, including the creation of bounding boxes and questions, using the open-call task on the Toloka crowdsourcing platform. The annotations were generated from scratch, utilizing exclusively the CC~BY-licensed images from MS~COCO. The annotators were asked to select the images containing the objects they found subjectively interesting and then compose questions about these objects. Then, for each question-image pair, we asked the annotators to select the answer on the image using a bounding box, allowing us to exclude unanswerable questions. Although it was possible to facilitate the question composition using such models as DH-GAN \citep{Kai:21}, we decided to stick to the pure crowdsourcing approach. This also allowed us to avoid synthetic data in our task and acquire the more natural formulations made by real humans.

We adopted the well-known methodology called Find-Fix-Verify \citep{Bernstein:10} for crowdsourcing creative datasets. It separates content production and verification tasks, and enables solving both tasks using crowdsourcing at scale. Our experience shows that the key issue in creative tasks is to ensure that \emph{all the annotators understand the task the same way as we do}. Thus, we had to run multiple iterations of task design that finally resulted in seven-stage annotation pipeline in Figure~\ref{fig:pipeline}. We will describe these stages in four coherent parts in the subsequent subsections. At the verification tasks the annotator who submitted the bounding box or a question and the annotator who verified it were two different people.

\subsection{\label{sub:annotators}Annotator Selection}

As composing good questions and drawing good bounding boxes imply a fair amount of creativity, we had to select the annotators who understand and feel the task the same way as we do. Besides this requirement, we needed the annotators to be able to actually solve it~--- by being able to formulate grammatically correct questions in English. As a result, we designed a two-step admission procedure for annotator selection that included a language test and a question verification task.

\paragraph{Language Test.} We had to make sure they have good reading and writing skills in English. Thus, we designed a single-choice test of five questions that was similar to the reading comprehension part of English exams. Each question contained a paragraph of approximately ten complex English sentences and was provided with four possible interpretations. Only one interpretation was correct. Since during prototyping we found that failing this task led to further problems with the question composition, we required the annotators to solve the test without any mistakes.

\paragraph{Question Verification.} After the annotators passed the language test, we wanted them to get a good understanding of what we expect them to produce. During prototyping we found that question composition part required additional attention, so the annotators had to solve the same verification task as described in Section~\ref{sub:questions}: given an image, a bounding box, and a question, confirm whether the question is well-formulated according to our strict requirements. We manually annotated the qualification dataset for this task. Those who passed this task were admitted for the real annotation.

\subsection{\label{sub:boxes}Bounding Boxes}

After sampling images from MS~COCO and selecting the right annotators for our task, we performed \emph{bounding box annotation} in two steps. First, we asked the annotators to pick one large unique object in the given image and draw a tight bounding box around it. Then, for each pair of image and bounding box, we asked the same annotators to check the submitted bounding boxes against the same instruction that we showed in the previous step.

\subsection{\label{sub:questions}Question Composition}

After obtaining the bounding boxes, we performed the similar steps to produce questions about the selected objects. We found this part to be the most challenging in our entire annotation task and spent most of our pipeline development time on it. First, given an image and a bounding box, we asked the annotators to compose a simple question in plain English that will allow one to find the object selected in the bounding box. Then, we asked the annotators to check the submitted questions against the same instruction that we showed them before. Since we have only one bounding box per image, we composed only one question per image.

\subsection{\label{sub:processing}Post-Processing}

We performed three additional steps to ensure a high quality of our dataset to exclude poorly-formulated, leaking, and potentially offensive instances.

\paragraph{Unanswerable Questions.} After receiving the entire dataset, we decided to ask the annotators to perform the same ask as the algorithms should: given an image and a question, draw a bounding box around the answer. We were able to establish the crowdsourcing baseline for further use (see Section~\ref{sec:metrics} for more details). Also, we managed to exclude from our dataset the instances for which the ground truth bounding boxes were significantly diverging from the newly-annotated bounding boxes.

\paragraph{Intersection Avoidance.} Since we used images from MS~COCO, we explicitly checked the overlap between bounding boxes in our dataset and in the original dataset. About 20\% of them had non-empty overlap, so we put all such instances into the train dataset. Otherwise the dataset splits were random.

\paragraph{Offensive Content.} During the dataset inspection, we found certain unacceptable questions suggesting offensive content. There were two prominent examples. First, there was a question ``What can hit the animals?'' for an image showing two zebras in a clearly non-hostile environment with a rock on sand.\footnote{\url{https://toloka-cdn.azureedge.net/wsdmcup2023/000000535978.jpg}} We reformulated the question. Second, there was a photo of a zebra shot by a group of hunters with the corresponding question. We fixed this by keyword filtering.

\section{\label{sec:metrics}Metrics and Baselines}

In our task, the answers correspond to bounding box coordinates, with only one bounding box per image. Therefore, we employ the \emph{intersection over union} ($\operatorname{IoU}$), also known as the \emph{Jaccard index}, as our evaluation criterion. For the $i$-th image, we define it as follows:
\begin{equation*}
  \operatorname{IoU}_i = \frac{I_i}{U_i}\text{,}
\end{equation*}
where $I_i$ represents the intersection area between the ground truth bounding box and the predicted bounding box, and $U_i$ is the union of these boxes. Consequently, for the entire dataset of $N$ images, the evaluation criterion is the \emph{average intersection over union}, denoted as $\operatorname{AIoU}$:
\begin{equation*}
  \operatorname{AIoU} = \frac{1}{N} \sum^N_{i = 1} \operatorname{IoU}_i\text{.}
\end{equation*}

\noindent{}For convenience, we multiply the $\operatorname{IoU}$ values by $100$. We used the following baselines to estimate the human and machine performance on our task. We additionally reported the prediction accuracy values that were obtained by choosing the threshold value for $\operatorname{IoU}$ and treating the instances for which the threshold was passed as the correct ones. We used two common thresholds, $\operatorname{IoU} = 50$ and $\operatorname{IoU} = 70$, and denote the accuracy values as $\operatorname{IoU} > 50$ and $\operatorname{IoU} > 70$, correspondingly.

\paragraph{Crowdsourcing.} We evaluated how well non-expert human annotators can solve our task by running a dedicated round of crowdsourcing annotations on Toloka. We found them to tackle this task successfully without knowing the ground truth. On all three subsets of our data, the average $\operatorname{IoU}$ value is $87.124 \pm 0.746$, which we consider as a \emph{strong human baseline} for our task. Krippendorff's $\alpha$ coefficients for the public test is $0.68$ and for the private test is $0.66$, showing the decent agreement between the responses; we used $1 - \operatorname{IoU}$ as the distance metric when calculating the $\alpha$ value.

\paragraph{OFA + SAM.} The first baseline is zero-shot and is primarily based on OFA \citep{Wang:22}, combined with bounding box correction using SAM. To solve the task, we followed a two-step zero-shot setup. First, we addressed visual question answering, where the model was given a prompt ``\{question\} Name an object in the picture'' along with an image. The model provided the name of a clue object to the question. In the second step, an object corresponding to the answer from the previous step was annotated using the prompt ``which region does the text "\{answer\}" describe?'', resulting in $\operatorname{IoU} = 42.462$. Subsequently, with the obtained bounding boxes, SAM generated the corresponding masks for the annotated object, which were then transformed into bounding boxes. This enabled us to achieve $\operatorname{IoU} = 44.851$ with this baseline.

\paragraph{OFA + SAM (VQA without Image).} We added the ablation study that answers questions without images as a new baseline model called OFA + SAM (VQA without Image). In particular, this method performed visual question answering by asking the question to the OFA model with blank white image and then drawing a bounding box corresponding to the obtained textual answer on the original image. This ablation shows that image is important for answering questions. The results were worse than the original OFA + SAM baseline, demonstrating $\operatorname{IoU} = 39.075$ vs. $\operatorname{IoU} = 44.851$ on the private test subset of our dataset.

\paragraph{OVSeg + SAM.} Another zero-shot baseline, called OVSeg \citep{Liang:23}, utilizes SAM \citep{Kirillov:23} as a proposal generator instead of MaskFormer in the original setup. This approach achieved $\operatorname{IoU} = 35.073$ on the private test subset.

\paragraph{Kosmos-2.} We also evaluated a grounding multi-modal large language model Kosmos-2 \citep{Kosmos-2} in a zero-shot setup. We addressed the visual grounding task with ``Find an object which answers the question. Question: "\{question\}". Answer:''. This baseline demonstrated $\operatorname{IoU} = 22.571$ on private test.

\paragraph{YOLOR + CLIP.} Our last baseline used a detection model, YOLOR \citep{Wang:21}, to generate candidate rectangles. Then, we applied CLIP \citep{Radford:21} to measure the similarity between the question and a part of the image bounded by each candidate rectangle. To make a prediction, it used the candidate with the highest similarity. This baseline method achieved $\operatorname{IoU} = 21.292$ on the private test subset.

\section{\label{sec:evaluation}Evaluation}

To assess our dataset beyond zero-shot baselines and crowd annotators, we conducted a large-scale open-call competition at WSDM~Cup, which took place alongside the WSDM '23 conference.\footnote{\url{http://www.wsdm-conference.org/2023/program/wsdm-cup}} To ensure fair participation, we hosted the competition on CodaLab. Participants were given access to the complete \emph{train} subset for training their models, as well as a masked \emph{public test} subset for the leaderboard competition.\footnote{\url{https://codalab.lisn.upsaclay.fr/competitions/7434}} The final rankings were determined based on performance on a concealed \emph{private test} subset, utilizing Docker images provided by the participants and executed on our servers. The inference code was required to complete within one hour on an Azure virtual machine equipped with 16~CPU cores, 200~GB of RAM, and one NVIDIA A100 80~GB GPU. We received a total of 48 participants in our competition, of which 9 submitted their code for the final stage. For the sake of brevity, Table~\ref{tab:results} reports only top-3 performance along with the above-described baselines. We put a more complete table with the competition results to supplementary materials. In the following paragraphs, we provide a brief overview of the methodologies employed by the three winning teams during the reproduction phase on the private test subset of our dataset.

\begin{table}[t]
\centering
\caption{\label{tab:results}Baselines and final top-3 team standings on the \emph{private test} subset, obtained at the reproduction phase of our competition; for visual convenience, we multiplied the IoU values by 100; out of 48 participants, only 9 submitted their code during the reproduction phase. Baseline methods did not participate in the competition, their places are denoted as ``---''.}
\begin{tabular}{ccrrr}\toprule
\textbf{Place} & \textbf{Team} & \textbf{IoU} & \textbf{IoU > 50} & \textbf{IoU > 70} \\\midrule
\color{gray}--- & \color{gray}Crowdsourcing & \color{gray}$87.154$ & \color{gray}$0.954$ & \color{gray}$0.914$\\
1 & \ttfamily{}wztxy89                  & $\mathbf{76.347}$ & $\mathbf{0.834}$ & $\mathbf{0.786}$\\
2 & \ttfamily{}jinx, Zhouyang\_Chi      & $\mathbf{76.342}$ & $\mathbf{0.841}$ & $\mathbf{0.779}$\\
3 & \ttfamily{}komleva.ep               & $\mathbf{75.591}$ & $\mathbf{0.756}$ & $\mathbf{0.772}$\\
\color{gray}--- & \color{gray}OFA + SAM & \color{gray}$44.851$ & \color{gray}$0.470$ & \color{gray}$0.431$\\
\color{gray}--- & \color{gray}OFA + SAM (VQA without Image) & \color{gray}$39.075$ & \color{gray}$0.407$ & \color{gray}$0.370$\\
\color{gray}--- & \color{gray}OVSeg + SAM & \color{gray}$35.073$ & \color{gray}$0.372$ & \color{gray}$0.309$\\
\color{gray}--- & \color{gray}Kosmos-2 & \color{gray}$22.571$ & \color{gray}$0.243$ & \color{gray}$0.073$\\
\color{gray}--- & \color{gray}YOLOR + CLIP & \color{gray}$21.292$ & \color{gray}$0.209$ & \color{gray}$0.201$\\\bottomrule
\end{tabular}
\end{table}

\begin{description}
  \item[3\textsuperscript{rd} Place.] The only single-person winning team, \texttt{komleva.ep}, fine-tuned the pre-trained multi-modal OFA model \citep{Wang:22} on the competition dataset. In order to increase the prediction quality, this team additionally used data from the pre-processed GQA dataset \citep{Hudson:19}.
  \item[2\textsuperscript{nd} Place.] The team \texttt{jinx, Zhouyang\_Chi} devised a three-step pipeline solution. First, at the \emph{coarse tuning} step, they generated textual pseudo answers for the questions and tuned the OFA model to produce textual answers. Then, at the \emph{fine tuning} step, they used prompt engineering of the coarse-tuned OFA model to draw the bounding boxes. Finally, at the \emph{post-processing} step, they ran an ensemble of these coarse- and fine-tuned models to propose and select the best bounding box candidate.
  \item[1\textsuperscript{st} Place.] The \texttt{wztxy89} team created a variant detector using Uni-Perceiver as the multi-modal backbone network \citep{Zhu:22}, with ViT-Adapter for cross-modal localization \citep{Chen:23}, and DINO as the prediction head \citep{Zhang:23}. They also included an auxiliary loss \citep{Kirillov:19} and a test-time augmentation module for improved performance, which helped them win the challenge \citep{gao2023champion}.
\end{description}

Even though the winning systems significantly outperformed our machine learning baselines, no system approached the non-expert level of human performance in our task.

\section{\label{sec:errors}Error Analysis}

\begin{table}[t]
\centering
\caption{\label{tab:errors}Distribution of error types of the crowdsourcing baseline and top-3 models from our challenge as described in Section~\ref{sec:evaluation}. The analysis is based on a random sample of 100 \emph{most challenging images}.}
\begin{tabular}{lrrrr}\toprule
\textbf{Error Class} & \textbf{Crowdsourcing} & \textbf{1\textsuperscript{st} Place} & \textbf{2\textsuperscript{nd} Place} & \textbf{3\textsuperscript{rd} Place} \\\midrule
Small Object & 26 & 13 & 13 & 11 \\
Insignificant Error & 19 & 13 & 12 & 16 \\
Wrong Object Predicted & 5 & 24 & 24 & 20 \\
Inaccurate Ground Truth & 11 & 9 & 9 & 7 \\
Inaccurate Prediction & 7 & 6 & 7 & 8 \\
Wrong Question, Correct Prediction & 20 & 12 & 13 & 14 \\
Wrong Question, Incorrect Prediction & 0 & 8 & 7 & 8 \\
Question Ambiguity & 12 & 15 & 15 & 16 \\
\bottomrule
\end{tabular}
\end{table}

\begin{figure}[t]
  \centering
  \subcaptionbox{What helps to tie hair? \textbf{Small Object.}}{\includegraphics[width=.235\linewidth]{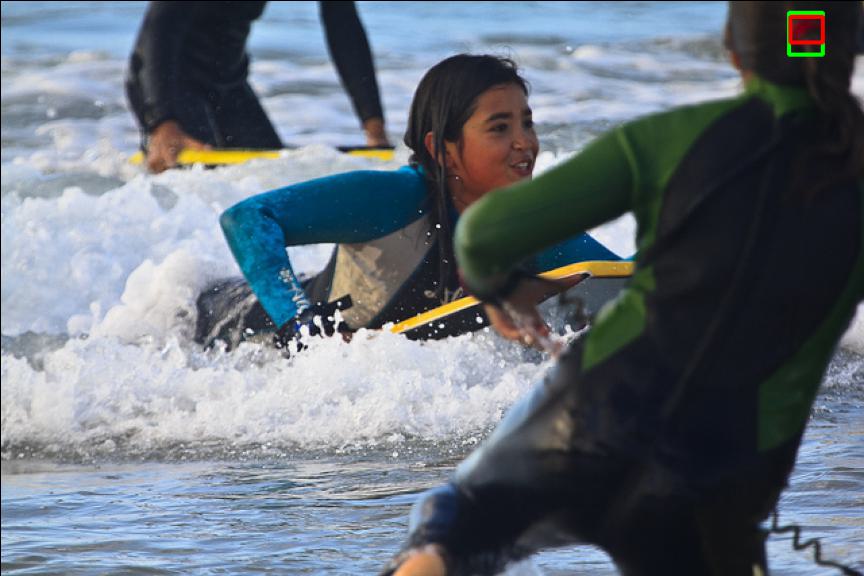}}
  \hfill
  \subcaptionbox{What is a device for giving light, one consisting of an electric bulb together with its holder ? \textbf{Insignificant Error.}\medskip}{\includegraphics[width=.235\linewidth]{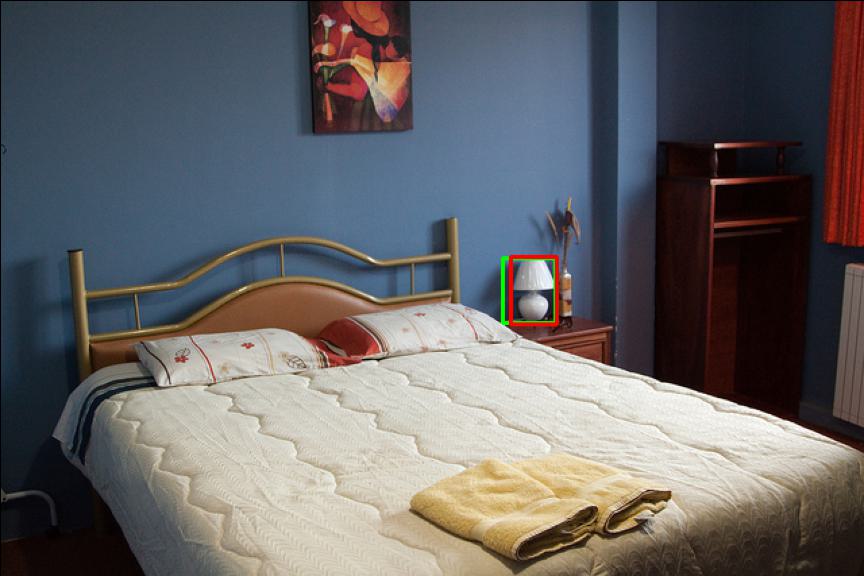}}
  \hfill
  \subcaptionbox{Which beverage is made of grapes? \textbf{Wrong Object Prediction.}}{\includegraphics[width=.235\linewidth]{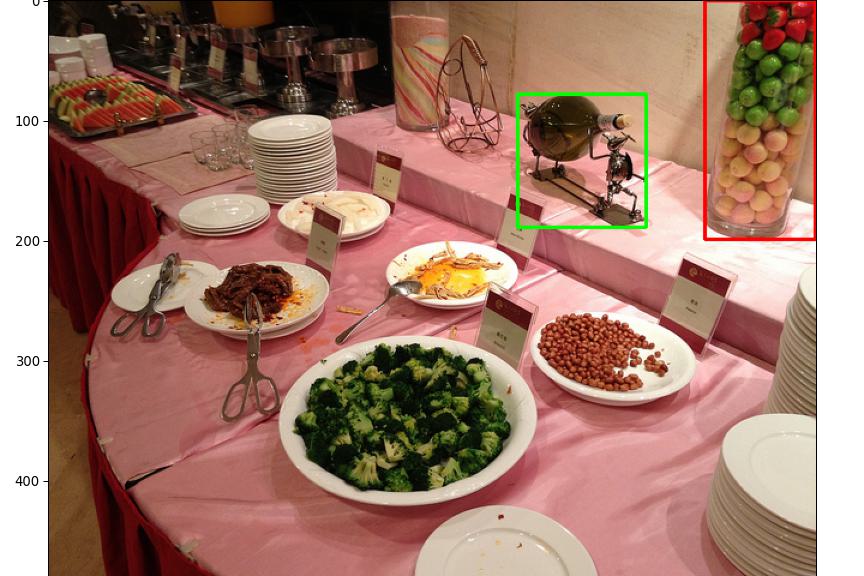}}
  \hfill
  \subcaptionbox{What do you play with on a windy day? \textbf{Inaccurate Ground Truth.}}{\includegraphics[width=.235\linewidth]{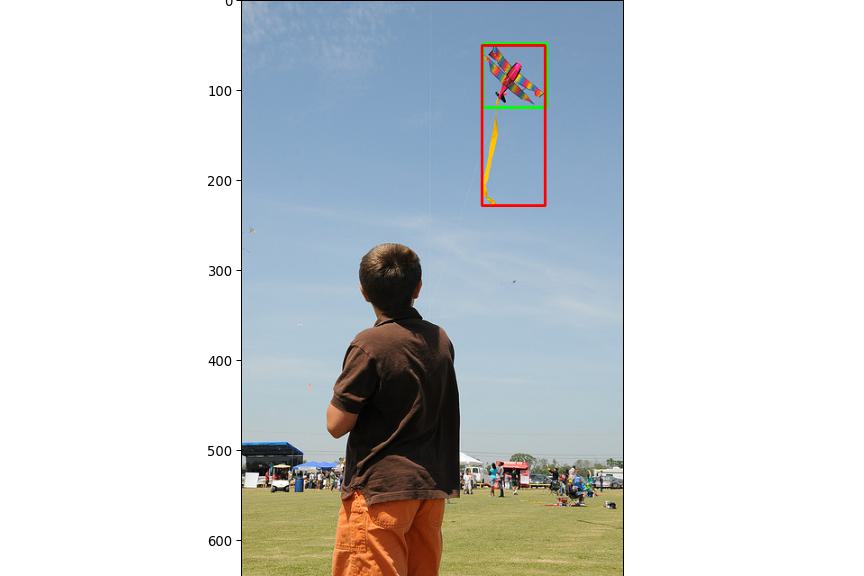}}
  \hfill
  \subcaptionbox{What is used to direct the traffic? \textbf{Inaccurate Prediction.}}{\includegraphics[width=.235\linewidth]{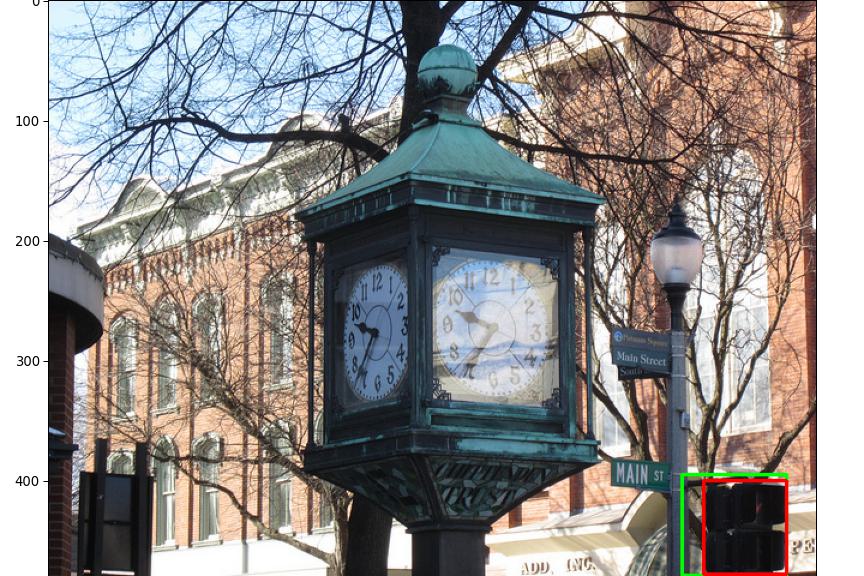}}
  \hfill
  \subcaptionbox{What can I open to let fresh air into the room? \textbf{Wrong Question, Correct Prediction.}}{\includegraphics[width=.235\linewidth]{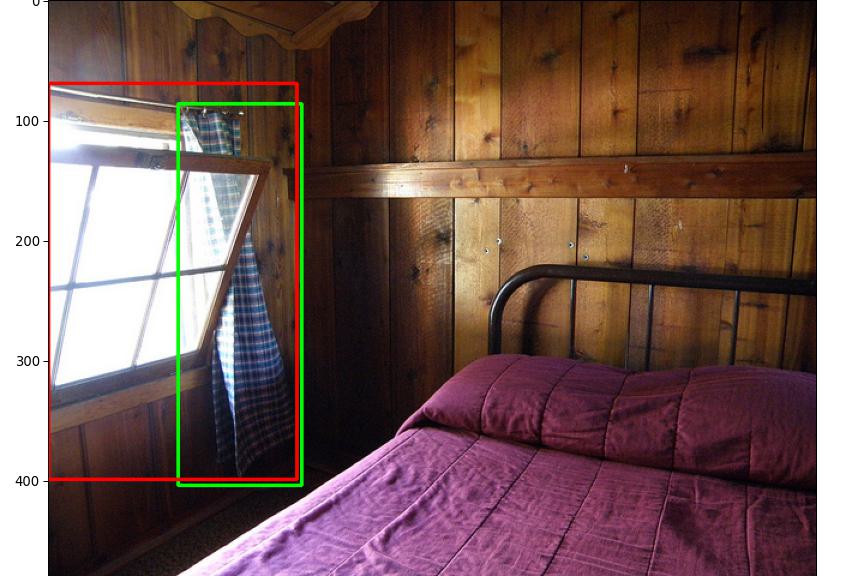}}
  \hfill
  \subcaptionbox{Where can I get supplies? \textbf{Wrong Question, Incorrect Prediction.}}{\includegraphics[width=.235\linewidth]{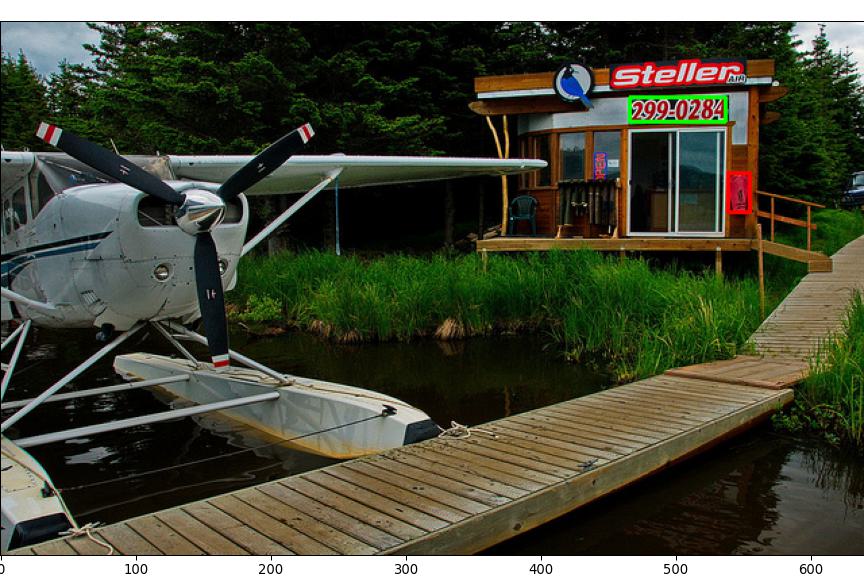}}
  \hfill
  \subcaptionbox{What can we wear? \textbf{Question Ambiguity.}}{\includegraphics[width=.235\linewidth]{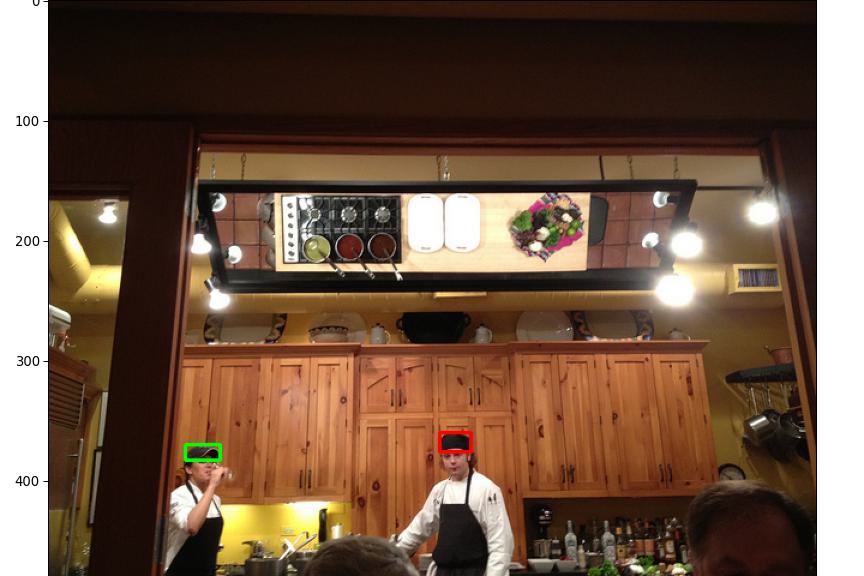}}
  \caption{\label{fig:errors}Typical examples of error classes we observed during the analysis. Image captions include the associated questions followed by the error class labels and are provided. In the images, \textcolor{red}{predictions are red-colored} and \textcolor{OliveGreen}{ground truth is green-colored}.}
\end{figure}

Before performing error analysis, we evaluated whether the errors made by three top-performing systems were similar or not. We took their predictions on the \emph{private test} subset and then computed the Krippendorff's $\alpha$ coefficient similarly to Section~\ref{sec:metrics}. The coefficient value of $0.77$ demonstrated a moderate agreement between responses of all three systems. As this indicates that all the three systems tend to produce similar correct and incorrect responses, we further analyzed the errors made by these systems and our crowdsourcing baseline as reported in Section~\ref{sec:evaluation}.

First, we sampled the data instances from the private test set where $\operatorname{IoU}$ of humans and models was less than $80$, which resulted in 355 out of 4,504 instances (approximately 8\%). \textbf{This sample is heavy biased towards the most challenging instances} as all the models showed poor results on them; it is \emph{not representative of the entire dataset}. Out of these 355, we sampled 100 instances and manually evaluated the quality of the obtained bounding boxes. This provided us with 100 judgements per method. We identified the following eight error classes as summarized in Table~\ref{tab:errors} and Figure~\ref{fig:errors}:
\begin{itemize}\itemsep0em
  \item \textbf{Small Object.} The target object is very small, thus it was difficult to draw a bounding box precisely because of the low resolution.
  \item \textbf{Insignificant Error.} The difference between prediction and ground truth is marginal (roughly, it means that $\operatorname{IoU}$ was greater than $70$).
  \item \textbf{Wrong Object Predicted.} The prediction is entirely incorrect, so $\operatorname{IoU}$ was zero.
  \item \textbf{Inaccurate Ground Truth.} The prediction is more accurate than ground truth.
  \item \textbf{Inaccurate Prediction.} The prediction is significantly less accurate than ground truth.
  \item \textbf{Wrong Question, Correct Prediction.} The ground truth does not answer the question but the corresponding prediction does.
  \item \textbf{Wrong Question, Incorrect Prediction.} Neither the ground truth nor the prediction answer the question.
  \item \textbf{Question Ambiguity.} The question might not have a single correct answer (for example, when there are several objects answering the same question). Also, we notice a few cases when the question is unclear or ambiguous (e.g., ``What is he doing'').
\end{itemize}

Having compared the predictions of the crowd annotators and top-performing machine learning systems, we noticed three facts. First, \emph{it was especially difficult for crowd to draw a bounding box around small objects} (the number of such errors is much higher than for models). Second, \emph{the models make a completely wrong prediction more often than the crowd}. However, even when the prediction is incorrect, the models still predict some object on the image and not just a random bounding box. Third, \emph{a significant amount of crowd errors is caused by the wrong or ambiguous question}. In most of such cases the crowd gives a correct answer. This is inevitable in crowdsourced datasets even with rigorous quality control approach like we used. It is worth noting that this observation is only applicable to the cases where all four approaches fail to make a good prediction (only $8\%$ of the data).

\section{\label{sec:conclusion}Conclusion}

In our grounding visual question answering task, the inputs consisted of an image and a question, with the output being the corresponding bounding box. While the top-performing systems showed remarkable improvement compared to the baselines, none of them surpassed non-expert annotators by a significant margin. We consider this fact important, as it indicates that our benchmark remains relevant until larger multi-modal models become accessible. The entire dataset, except for the images themselves, was generated through crowdsourcing on the Toloka platform, rendering it a valuable resource for the creation of demanding benchmarks. As a future work, we consider increasing the dataset size and adjusting our annotation pipeline to produce questions on different regions in the same image.

We foresee the following potential \textbf{downstream applications} of our dataset and derivative models beside the evaluation of machine learning models:
\begin{itemize}\itemsep0em
  \item \textbf{Visual Search.} With accurate bounding boxes, grounding VQA enables better understanding and recognition of objects in images, allowing users of e-commerce platforms to query products based on their appearance rather than relying only on text-based search.
  \item \textbf{Augmented Reality (AR).} Accurate bounding box annotation can help in integrating virtual objects into real-world scenes during AR applications. Grounding VQA aids in object recognition and aligning virtual content with real-world objects, and can assist in image annotation, making it easier to search images based on specific object queries.
  \item \textbf{Robotics.} For robots to interact with objects in their environment effectively, accurate object localization is crucial. grounding VQA can be utilized to identify and track objects, enabling robots to navigate, grasp, and manipulate objects in a more intelligent and precise manner.
\end{itemize}

However, our dataset has the following \textbf{limitations}:
\begin{itemize}\itemsep0em
  \item \textbf{Dataset Bias.} We only consider images from MS~COCO, which itself may contain biases related to gender and race. Additionally, the questions and selected objects chosen by annotators may introduce bias or have limited variability, potentially limiting the generalizability of models trained on this dataset.
  \item \textbf{English-Only Questions.} This dataset focuses solely on English questions. However, this narrow focus may restrict the ability of models to handle other languages and cultures.
  \item \textbf{Real-World Applications.} Since there are no production-level applications of the proposed grounding VQA setup yet, future real-world applications may involve more complex questions that require deeper understanding, reasoning, and context awareness.
\end{itemize}

We wish to highlight certain \textbf{potential negative social impacts} of models trained on our dataset:

\begin{itemize}\itemsep0em
  \item \textbf{Reinforcing Bias.} Due to potential biases present in the dataset, models trained on it may inadvertently perpetuate societal inequalities and discrimination when deployed in real-world applications.
  \item \textbf{Ethical Use of Models.} As grounding VQA models become more advanced, they can be exploited for malicious purposes, such as privacy invasion. It is crucial to establish proper safeguards and guidelines to prevent misuse and protect individuals' rights and well-being.
\end{itemize}

\paragraph{Acknowledgements.} This work would not have been possible without the collaborative efforts of people from different teams in Toloka. Individuals listed in each contribution role are arranged alphabetically based on their surnames. We thank Ekaterina Fedorenko, Natalia Fedorova, Ujwal Gadiraju, Valentina Mikhno, and Evgeniya Sukhodolskaya for helping in organizing our competition at WSDM~Cup. We acknowledge the invaluable efforts of Oleg Pavlov, Mikhail Potalitsyn, and Rosmiyana Shekhovtsova, whose expertise was crucial for the annotation pipeline design. We express gratitude to Anastasia Egupova, Aleksei Gerasimenko, Timur Pevzner, Kuen Pham, Ekaterina Saenko, and Anna Stepanova for their help in raising awareness of the competition among the community, allowing to attract 48 participants from across the world. We are grateful to Egor Babkin, Dmitry Lekomtsev, and Andrei Voitovich for building the Web presence of our competition. We acknowledge the invaluable contribution of Tatiana Ignatova, Daria Kalakina, and Victoriya Vidma in navigating complex legal and financial aspects. Last but not least, we would like to thank the CodaLab and the WSDM Cup teams, especially Hady W.\ Lauw, and the competition participants, for making it a big success.

\bibliographystyle{plain}
\bibliography{tolokavqa}

\clearpage

\appendix

\section{Competition Results at WSDM~Cup 2023}

The competition website was located at \url{https://toloka.ai/challenges/wsdm2023/}, while the leaderboard and scoring were performed on the CodaLab platform: \url{https://codalab.lisn.upsaclay.fr/competitions/7434}. Before the start of the competition, all the parts of our dataset were frozen and did not change during the competition.

Our competition had three key phases: the practice phase, the evaluation phase, and the reproduction phase (Table~\ref{tab:timeline}). In September, we started the \emph{practice} phase to let the contestants get used to the task and training data, including the ground truth data. Then, we started the \emph{evaluation} phase using the public test dataset without ground truth labels. The contestants had to submit their predictions to the competition platform, which resulted in leaderboard updates. Finally, for the sake of reproducibility, soon after the end of the evaluation phase, we started the \emph{reproduction} phase. In this phase, we asked the contestants to provide their solution as a container image. We ran their code to obtain answers for the private test dataset to determine the winners.

Table~\ref{tab:fullresults} shows the competition results. Even though the participants managed to improve dramatically upon our baselines, none of the participating systems outperformed our crowdsourcing baseline.

\begin{table}[htbp]
\centering
\caption{\label{tab:timeline}Complete timeline of our competition at WSDM Cup 2023}
\begin{tabular}{lr}\toprule
\textbf{Event} & \textbf{Date} \\\midrule
Practice Starts & September 16, 2022 \\
Evaluation Starts & September 30, 2022 \\
Evaluation Ends & December 16, 2022 \\
Reproduction Starts & December 19, 2022 \\
Reproduction Ends & January 16, 2023 \\
Post-Competition Starts & January 16, 2023 \\
WSDM Cup Workshops & March 3, 2023 \\\bottomrule
\end{tabular}
\end{table}

\begin{table}[htbp]
\centering
\caption{\label{tab:fullresults}Baselines and final team standings on the \emph{private test} subset, obtained at the reproduction phase of our competition; for visual convenience, we multiplied the IoU values by 100; out of 48 participants, only 9 submitted their code during the reproduction phase. Baseline methods did not participate in the competition, their places are denoted as ``---''.}
\begin{tabular}{ccc}\toprule
\textbf{Place} & \textbf{Team} & \textbf{IoU} \\\midrule
\color{gray}--- & \color{gray}Crowdsourcing & \color{gray}$87.154$ \\
1 & \ttfamily{}wztxy89                  & $\mathbf{76.347}$ \\
2 & \ttfamily{}jinx, Zhouyang\_Chi      & $\mathbf{76.342}$ \\
3 & \ttfamily{}komleva.ep               & $\mathbf{75.591}$ \\
4 & \ttfamily{}xexanoth                 & $74.667$ \\
5 & \ttfamily{}Man\_of\_the\_year       & $72.768$ \\
6 & \ttfamily{}Haoyu\_Zhang, KhylonWong & $71.998$ \\
7 & \ttfamily{}nika-li                  & $70.525$ \\
8 & \ttfamily{}blinoff                  & $62.037$ \\
9 & \ttfamily{}Ndhuynh                  & $61.247$ \\
\color{gray}--- & \color{gray}OFA + SAM & \color{gray}$44.851$ \\
\color{gray}--- & \color{gray}OFA + SAM (without Image) & \color{gray}$39.075$ \\
\color{gray}--- & \color{gray}OVSeg + SAM & \color{gray}$35.073$ \\
\color{gray}--- & \color{gray}Kosmos-2 &
\color{gray}$22.571$ \\
\color{gray}--- & \color{gray}YOLOR + CLIP & \color{gray}$21.292$ \\\bottomrule
\end{tabular}
\end{table}

\clearpage

\section{Annotator Instructions and Interfaces}

\begin{figure}[htbp]
  \centering
  \includegraphics[width=\textwidth]{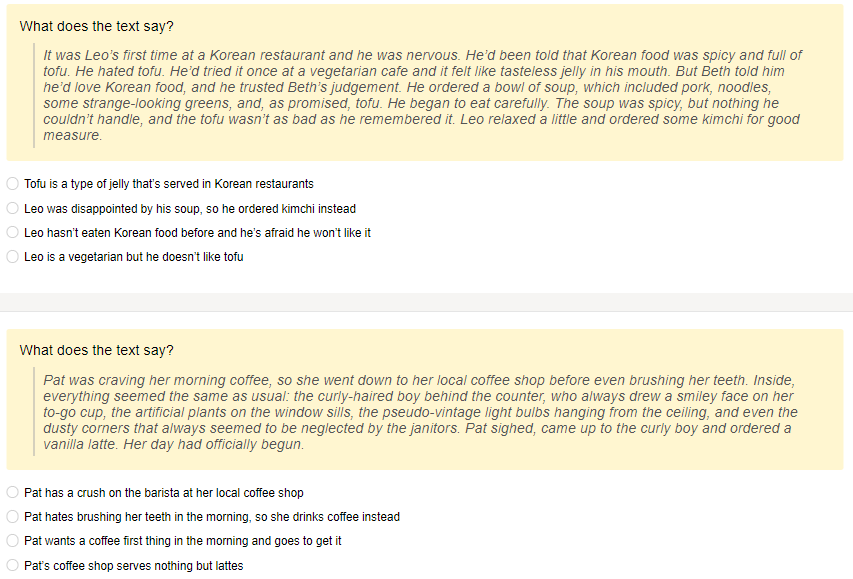}
  \caption{\label{fig:language-test}Screenshot of the interface of the language test assignment.}
\end{figure}

\begin{figure}[htbp]
  \centering
  \includegraphics[width=\textwidth]{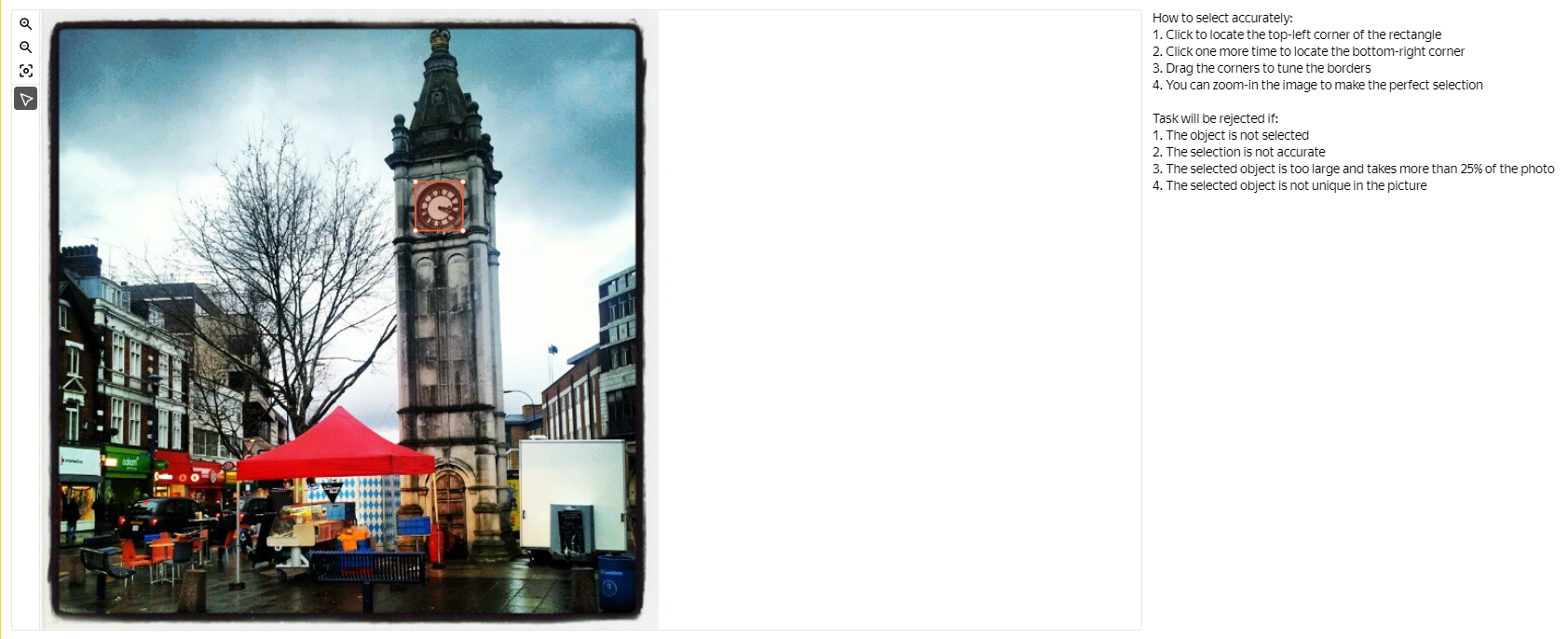}
  \caption{\label{fig:find-object}Screenshot of the ``Find an Interesting Object'' annotation interface.}
\end{figure}

\begin{figure}[htbp]
  \centering
  \includegraphics[width=\textwidth]{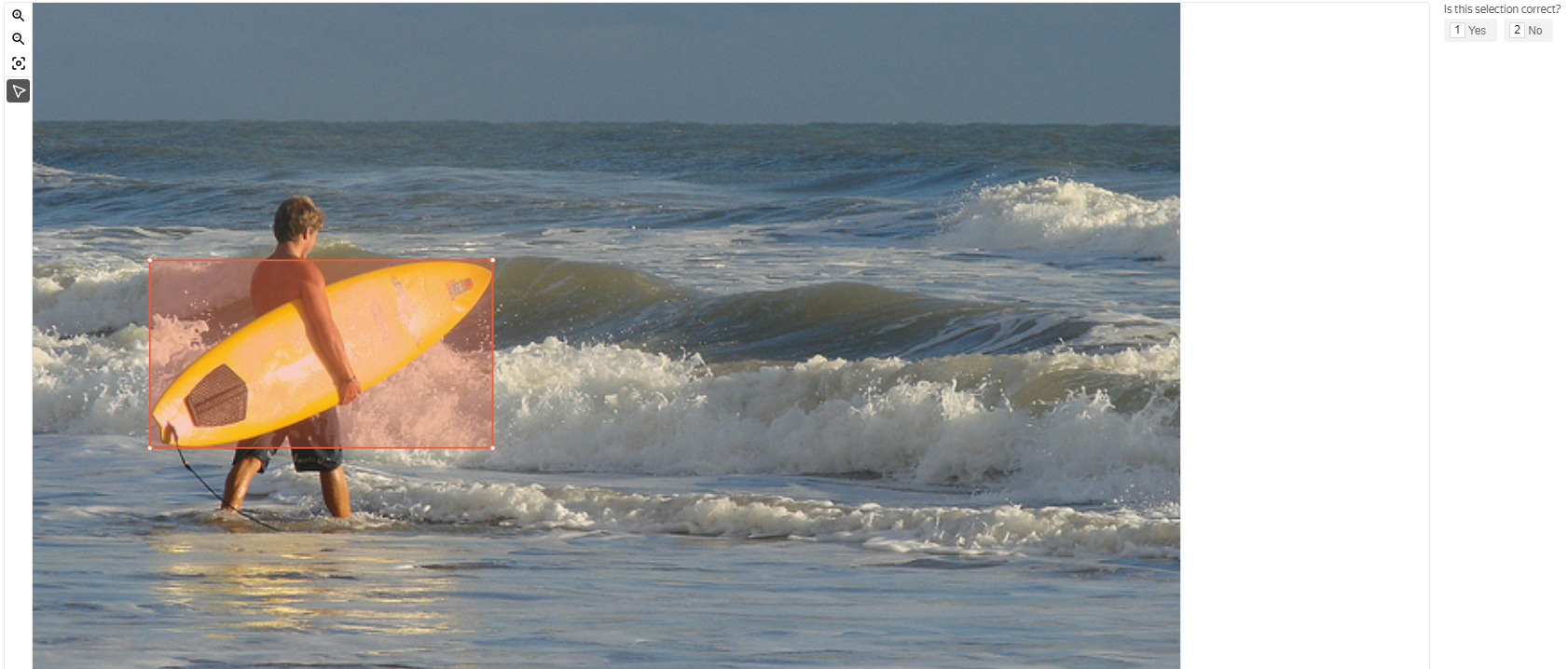}
  \caption{\label{fig:verify-object}Screenshot of the ``Verify Object Selection'' annotation interface.}
\end{figure}

\begin{figure}[htbp]
  \centering
  \includegraphics[width=\textwidth]{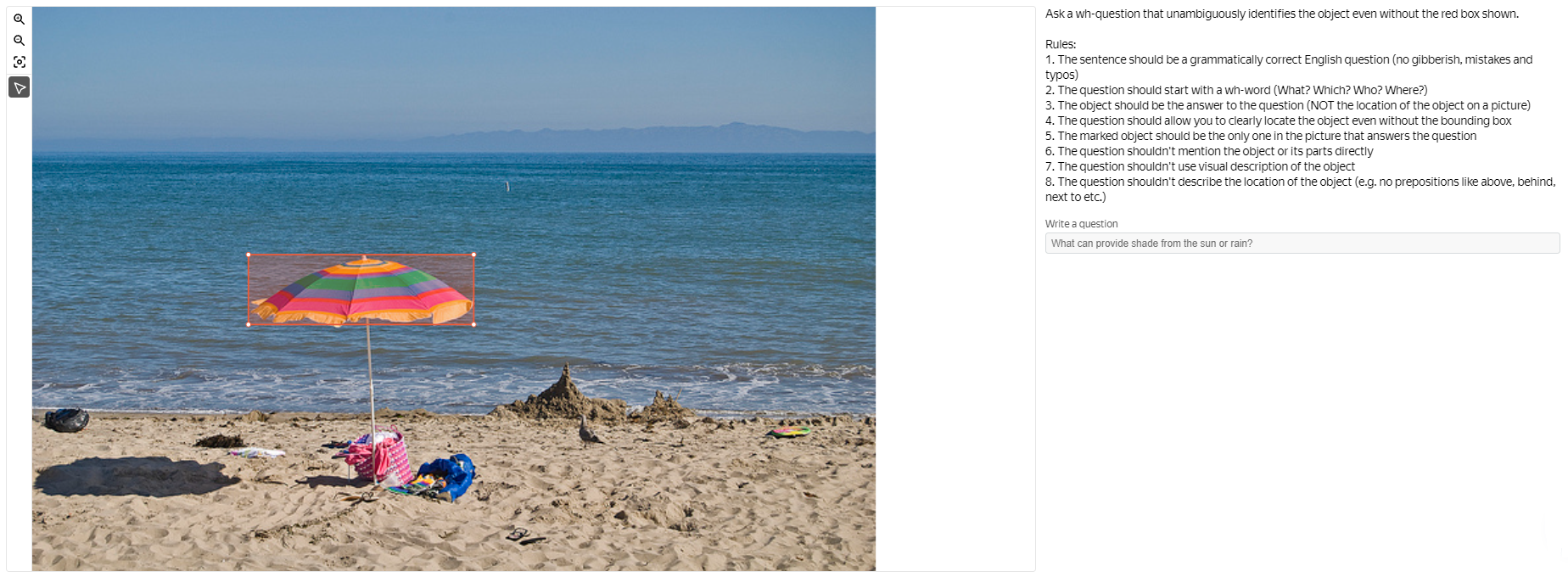}
  \caption{\label{fig:ask-question}Screenshot of the ``Ask a Question about the Object'' annotation interface.}
\end{figure}

\begin{figure}[htbp]
  \centering
  \includegraphics[width=\textwidth]{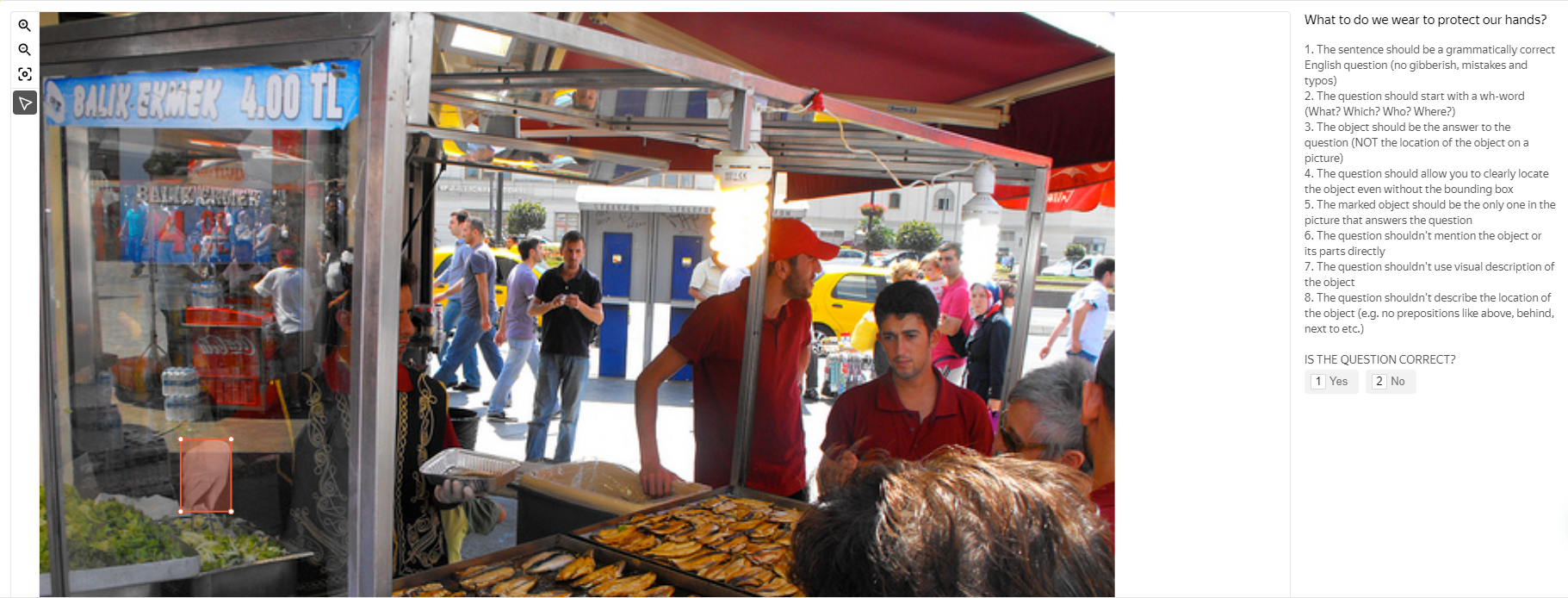}
  \caption{\label{fig:verify-question}Screenshot of the ``Verify Questions About Objects'' annotation interface.}
\end{figure}

Our annotation pipeline consisted of four phases, each corresponding to an annotation project: ``Find an Interesting Object'' for initial bounding box selection of objects in images, ``Verify Object Selection'' to confirm if the selected objects met the requirements, ``Ask a Question about the Object'' for generating questions about the selected objects, and ``Verify Questions About Objects'' to validate if the written questions met the requirements. We provide a screenshot of our language test in Figure~\ref{fig:language-test} and screenshots of our main annotation tasks in Figure~\ref{fig:find-object}, \ref{fig:verify-object}, \ref{fig:ask-question}, and \ref{fig:verify-question}.

\end{document}